\documentclass{article}

\usepackage{microtype}
\usepackage{graphicx}
\usepackage{subcaption}
\usepackage{booktabs} %

\usepackage{hyperref}

\usepackage[preprint]{icml2026}

\usepackage{amsmath}
\usepackage{amssymb}
\usepackage{mathtools}
\usepackage{amsthm}

\usepackage[capitalize,noabbrev]{cleveref}

\theoremstyle{plain}

\theoremstyle{definition}

\theoremstyle{remark}

\usepackage[textsize=tiny]{todonotes}

\usepackage{adjustbox}
\usepackage{wrapfig}
\usepackage{tcolorbox}
\usepackage{tabularx}
\usepackage{array}
\usepackage{booktabs}
\usepackage{makecell}
\usepackage{pifont} %
\usepackage{xspace}
\usepackage{multirow}
\newcommand{\cmark}{\textcolor{blue}{\ding{51}}} %
\newcommand{\xmark}{\textcolor{red}{\ding{55}}} %

\newcommand{\ourname}{$\infty$-\textsc{Thor}\xspace}

\icmltitlerunning{Beyond Needle(s) in the Embodied Haystack}

\begin{document}

\twocolumn[
  \icmltitle{Beyond Needle(s) in the Embodied Haystack: Environment, Architecture, \\ and Training Considerations for Long Context Reasoning}

  \icmlsetsymbol{equal}{*}

  \begin{icmlauthorlist}
    \icmlauthor{Bosung Kim}{yyy}
    \icmlauthor{Prithviraj Ammanabrolu}{yyy}
  \end{icmlauthorlist}

  \icmlaffiliation{yyy}{University of California, San Diego}

  \icmlcorrespondingauthor{Bosung Kim}{bosungkim@ucsd.edu}

  \icmlkeywords{Machine Learning, ICML}

  \vskip 0.3in
]

\printAffiliationsAndNotice{}  %

\begin{abstract}
We introduce \ourname, a new framework for long-horizon embodied tasks that advances long-context understanding in embodied AI.
\ourname provides:
(1) a generation framework for synthesizing scalable, reproducible, and unlimited long-horizon trajectories;
(2) a novel QA task, Needle(s) in the Embodied Haystack, where scattered clues across extended trajectories test agents’ long-context reasoning ability; and
(3) a long-horizon dataset and benchmark suite featuring complex tasks that span hundreds of environment steps, each paired with ground-truth action sequences.
To enable this capability, we explore architectural adaptations, including interleaved Goal-State-Action and Memory-Augmented Goal-State modeling,
alongside training strategies---context extension techniques and Context Parallelism---to equip VLM-based agents for extreme long-context reasoning and interaction.
Experiments highlight the challenges of our benchmark and provide insights into training strategies and model behaviors under long-horizon conditions.
Furthermore, we demonstrate sim-to-real transfer: our QA dataset alone improves photorealistic benchmark performance by up to +11.2\%, and we successfully integrate the framework with low-level manipulation controls.
Our work provides a foundation for the next generation of embodied AI systems capable of robust, long-term reasoning and planning.
The datasets and code can be found at \href{https://pearls-lab.github.io/infini-thor}{\texttt{pearls-lab.github.io/infini-thor}}.
\end{abstract}

\section{Introduction}

Real-world embodied reasoning is a sequential decision-making problem requiring long-horizon planning, where task success depends on both memorizing and reasoning over multiple events that occur far apart in time.
Using vision-language models (VLM) as policies for such tasks requires surpassing the key challenge of \textit{long-context} reasoning.
We seek to answer questions pertaining to what design choices matter in terms of environments, model architectures, and training methods when using VLMs for long-horizon embodied tasks.
To this end, we develop a new framework for long-horizon tasks designed to push the boundaries of long-context understanding in embodied AI.

\begin{figure*}[t]
\centering
\includegraphics[width=1.0\textwidth]{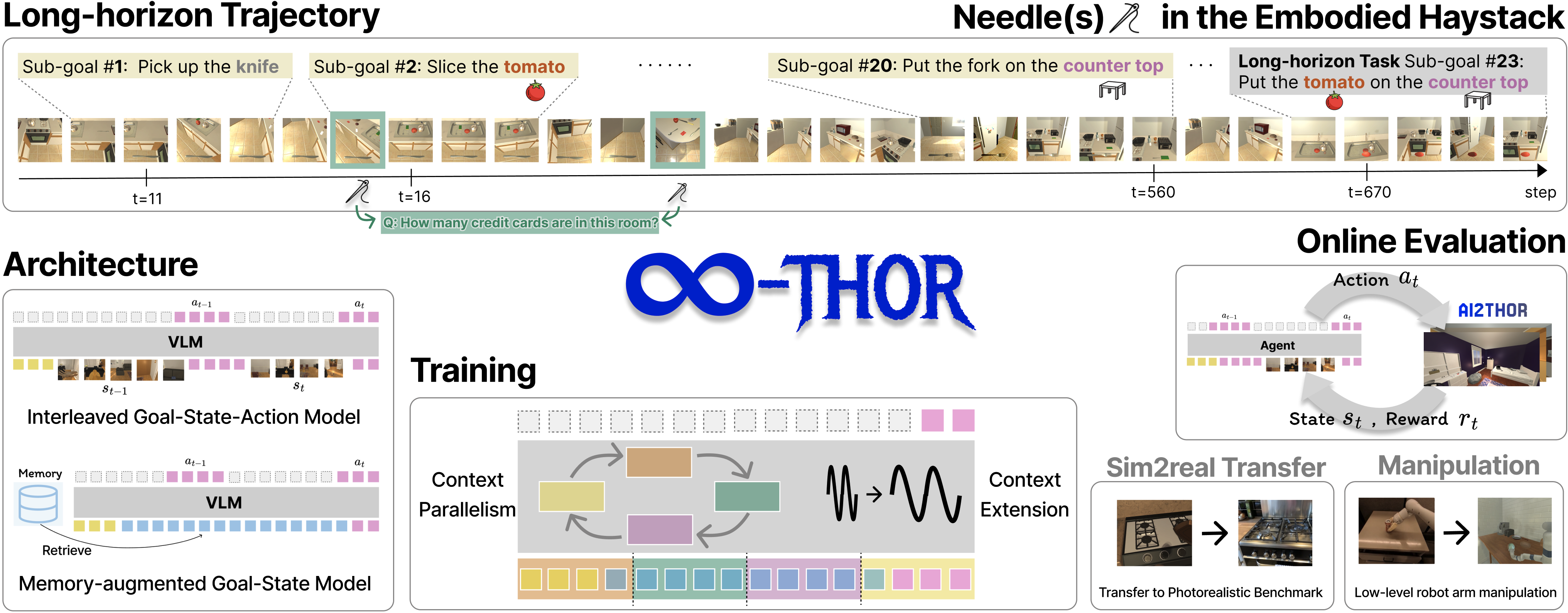}
\caption{
{Overview of \ourname.}
\textbf{Environment:} We introduce a framework for generating long-horizon trajectories, the \textit{Needle(s) in the Embodied Haystack} benchmark to test reasoning over scattered clues, and interactive online evaluation.
\textbf{Architecture:} We investigate VLM adaptations as policy models, including Interleaved Goal-State-Action and Memory-augmented models.
\textbf{Training:} We incorporate Context Parallelism and Context Extension for efficient long-context learning.
In experiments, we demonstrate Sim2Real transfer to photorealistic benchmark tasks and integration with low-level manipulation controls.
}
\vspace{-10pt}
\label{fig:fig1}
\end{figure*}

We introduce \ourname, a new framework for generation, training, and evaluation of long-horizon embodied tasks.
Our benchmark uniquely features tasks with a synthetic final goal, which involves multiple objects that appear at distant time steps, requiring multi-step reasoning across over hundreds of steps.
As illustrated in Figure \ref{fig:fig1}, an agent might observe a tomato at $t=16$ and a counter top much later at $t=560$; the final task given at $t=670$ requires the agent to retrieve the tomato and place it on the counter.
This setup highlights the challenge of long-horizon dependency, where key objects and locations must be remembered and acted upon after hundreds of steps. 

This long-horizon setup introduces a new challenging task, Needle(s) in the Embodied Haystack (NiEH).
Unlike the standard Needle in a Haystack task \citep{liu-etal-2024-lost}, which focuses on recalling a single clue in text, NiEH poses two main challenges: (1) multiple scattered clues (\textit{Needles}) and (2) multi-modal inputs that combine visual and linguistic observations from the environment (\textit{Embodiment}).
This task is designed to evaluate the agent's ability to recall and reason about previously encountered environmental details, such as identifying objects and recalling performed actions.

Going beyond static evaluations such as NiEH, \ourname also provides an interactive evaluation, allowing agents to execute policies and complete long-horizon tasks within a dynamic environment.
To support this, we release a trajectory dataset for training, with episodes over 400 steps in the training set and more than 600 steps in the dev and test sets.
These trajectories can be used for imitation learning, and our experiments show that access to longer context during training leads to significant performance gains, highlighting the importance of our dataset for long-context embodied reasoning.

With the environment established, we investigate various architectural considerations for embodied agents to operate under extreme sequence lengths.
We explore two complementary approaches: Interleaved Goal-State-Action model, which directly encodes the full perception–action history as a single multimodal sequence,
and Memory-Augmented Goal-State model, which replaces the historical trajectory with text summaries or retrieved visual memories to provide long-term contextual grounding.
Moreover, since standard VLMs are constrained by fixed context windows and cannot natively handle inputs exceeding 1M tokens, we evaluate long-context extension techniques, including linear interpolation, dynamic scaling, YaRN, and LongRoPE \citep{chen2023extendingcontextwindowlarge,ding2024longropeextendingllmcontext,peng2024yarn}.
Lastly, to support practical training and inference at these extended horizons, we incorporate Context Parallelism via Ring Attention \citep{liu2023ring}, enabling efficient fine-tuning on ultra-long sequences and further improving the model’s ability to reason over extended temporal contexts.

Finally, we provide comprehensive experiments and analyses, demonstrating both the challenges posed by our benchmark and the behavior of baseline models under long-horizon settings. 
Furthermore, we validate the real-world applicability of our framework through \textit{Sim2Real} transfer, showing that our QA dataset improves performance on photorealistic benchmarks.
We also demonstrate the framework's versatility by integrating it with ManipulaTHOR \cite{Ehsani2021ManipulaTHORAF} to enable low-level robot arm manipulation.

To summarize, our contributions are:
\textbf{(1)} We introduce \ourname, a new framework for generating, training, and evaluating long-horizon embodied tasks, featuring synthetic goals that require multi-step reasoning across hundreds of steps.
\textbf{(2)} We propose a novel embodied QA task, Needle(s) in the Embodied Haystack, requiring agents to recall and reason over multiple scattered clues across extended trajectories.
\textbf{(3)} We release a large-scale trajectory dataset and an interactive evaluation environment to support both offline imitation learning and online policy execution.
\textbf{(4)} We analyze architectural adaptations (Interleaved vs. Memory-Augmented) and training strategies (Context Extension, Context Parallelism) for handling long-context inputs.
\textbf{(5)} We present empirical results demonstrating the benefits of our framework, including successful Sim2Real transfer and integration with low-level manipulation controls.

\section{Related Work}

\begin{table*}[ht]
    \centering
    \small
    \caption{Comparison of benchmarks. We use Short ($<50$ steps), Medium ($50$-$300$ steps), and Long ($>300$ steps) to describe task horizon, reflecting the approximate number of environment steps required to complete a task in each benchmark. (Inter. w/ env: Interaction with the environment, Mod: Modality, GT: GT actions; single/multi in the QA set column denotes single- and multi-evidence question type. * indicates the number of annotations newly collected in that work.)
    }
    \label{tab:benchmark_comparison}
    \begin{tabular}{l|c|c|ccc|cc}
        \toprule
        \textbf{Benchmark / Platform} & \multicolumn{1}{c|}{\textbf{Task}} & \multicolumn{1}{c|}{\textbf{Inter.}} & \multicolumn{3}{c|}{\textbf{Dataset}} & \multicolumn{2}{c}{\textbf{QA set}} \\
        & \textbf{Horizon} & \textbf{w/ env} & {Mod} & {Avg steps} & {GT} & {single} & {multi} \\
        \midrule

        ProcTHOR \citep{deitke2022procthorlargescaleembodiedai} & \xmark & \cmark & \xmark & \xmark & \xmark & \xmark & \xmark \\
        MineDojo \citep{fan2022minedojobuildingopenendedembodied} & Long & \cmark & \xmark & \xmark & \xmark & \xmark & \xmark \\
        Habitat 3.0 \citep{puig2023habitat30cohabitathumans} & Long & \cmark & \xmark & \xmark & \xmark & \xmark & \xmark \\
        VirtualHome \citep{puig2018virtualhomesimulatinghouseholdactivities} & Short & \cmark & multi & 11.6 & \cmark & \xmark & \xmark \\
        EQA \citep{embodiedqa} & \xmark & \xmark & \xmark & \xmark & \xmark & \cmark & \xmark \\
        ALFRED \citep{shridhar2020alfred} & Medium & \cmark & multi & 50 & \cmark & \xmark & \xmark \\
        ALFWorld \citep{ALFWorld20} & Medium & \cmark & text & 50 & \cmark & \xmark & \xmark \\
        BEHAVIOR-100 \citep{srivastava2021behaviorbenchmarkeverydayhousehold} & Med/Long & \cmark & \xmark & \xmark & \xmark & \xmark & \xmark \\
        EAI \citep{li2024embodied} & Med/Long & \cmark & \cmark & 14.6* & \cmark & \xmark & \xmark \\
        BALROG \citep{paglieri2025balrogbenchmarkingagenticllm} & Long & \xmark & \xmark & \xmark & \xmark & \xmark & \xmark \\
        MM-EGO \citep{ye2025mmego} & \xmark & \xmark & \xmark & \xmark & \xmark & \cmark & \xmark \\
        \midrule 
        \textbf{\ourname} & $\infty$ & \cmark & multi & 627 & \cmark & \cmark & \cmark \\
        \bottomrule
    \end{tabular}
    \vspace{-10pt}
\end{table*}

\textbf{Long-horizon Planning in Virtual Environments.}
AI2THOR \citep{kolve2017ai2} provides interactive indoor environments widely used for embodied reasoning research, while ProcTHOR \citep{deitke2022procthorlargescaleembodiedai} extends these capabilities by procedurally generating scalable environments, potentially facilitating longer trajectories.
MineDojo \citep{fan2022minedojobuildingopenendedembodied} offers an open-ended platform within Minecraft, explicitly geared toward tasks requiring extensive long-term planning. 
Additionally, platforms such as VirtualHome \citep{puig2018virtualhomesimulatinghouseholdactivities} and Habitat 3.0 \citep{puig2023habitat30cohabitathumans} have demonstrated suitability for tasks involving long-term interactions and complex activity sequences.
However, all of these platforms only provide environments and do not include standardized datasets or benchmark suites to support training and evaluation for long-horizon embodied tasks.

\textbf{Embodied QA and Multimodal Needle in the Haystack Tasks.}
Embodied QA tasks, such as EmbodiedQA \citep{embodiedqa} and MM-EGO \citep{ye2025mmego}, require agents to answer questions grounded in visual observations with spatial and temporal reasoning, but without active environment interaction during evaluation.
Our NiEH task is also related to multimodal Needle in a Haystack (NiH) problems.
While early NiH focused on textual recall in long contexts \citep{liu-etal-2024-lost}, recent multimodal extensions add visual inputs \citep{wang2024needlemultimodalhaystack,wang-etal-2025-multimodal}, though they remain limited to shorter contexts (up to 72K tokens) and lack embodied reasoning or temporal dependencies.

\textbf{Datasets and Benchmarks for Long-horizon Embodied Tasks.}
Recent efforts have pushed toward long-horizon embodied tasks, where agents must complete multi-step goals with extended temporal dependencies.
While ALFRED \citep{shridhar2020alfred} and ALFWorld \citep{ALFWorld20} introduced multi-step instruction-following tasks with action annotations and textual grounding, their task horizons are relatively short, typically under 50 steps.
BEHAVIOR-100 \citep{srivastava2021behaviorbenchmarkeverydayhousehold} evaluates agent generalization on household activities, some of which require prolonged engagement, but mainly focus on single task.
BALROG \citep{paglieri2025balrogbenchmarkingagenticllm} is a benchmark for testing the agentic capabilities of long-context LLMs, but its scope is limited to games.
Recently, EAI \citep{li2024embodied} proposed a generalized interface to evaluate LLMs for embodied decision making, while our framework focuses on multimodal online evaluation with real-time rewards and a new long-horizon reasoning task (NiEH).

\textbf{Long-context  Benchmarks.}
Outside embodied AI, general benchmarks have addressed challenges in long-context reasoning.
Benchmarks, such as LongBench \citep{bai-etal-2024-longbench} and RULER \citep{hsieh2024rulerwhatsrealcontext}, focus on retrieval or summarization tasks.
GSM-$\infty$ \citep{zhou2025gsminfinitellmsbehaveinfinitely} extends GSM-8K \citep{cobbe2021trainingverifierssolvemath} to assess mathematical reasoning over extremely long textual inputs.
More recently, LMAct \citep{ruoss2025lmactbenchmarkincontextimitation} proposed a benchmark for evaluating frontier models' long-context multimodal decision-making on interactive game-based tasks, with up to 1M context lengths.

\section{\ourname: An Environment for Generating, Training, and Evaluating Long-horizon Embodied Tasks}

\begin{figure*}[t]
\centering
\subfloat[Needle in the Embodied Haystack: Single-evidence question types.]{
  \includegraphics[width=1.0\textwidth]{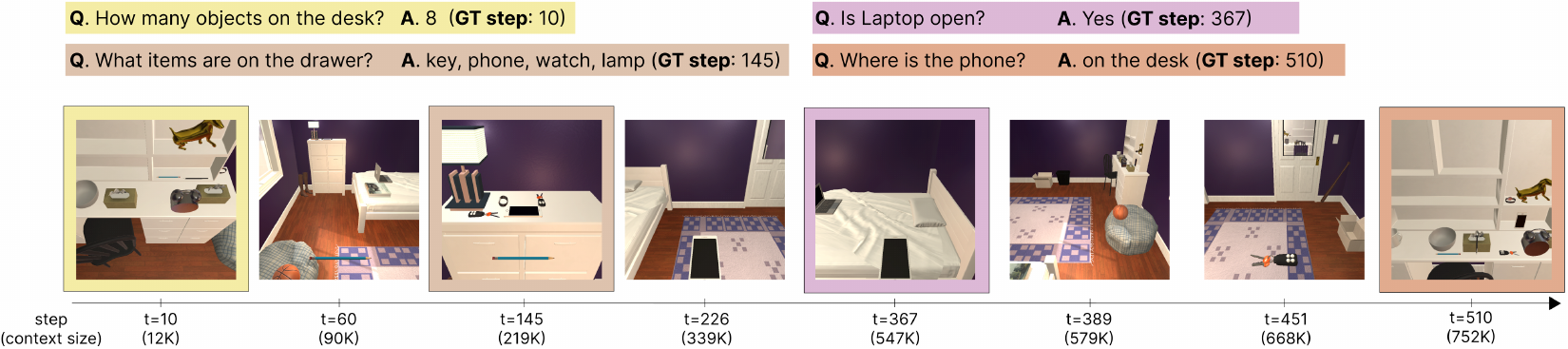}
  \label{fig:example_NiEH}
}
\vspace{0.2em} %
\subfloat[Needle\textbf{s} in the Embodied Haystack: Multi-evidence question types.]{
  \includegraphics[width=1.0\textwidth]{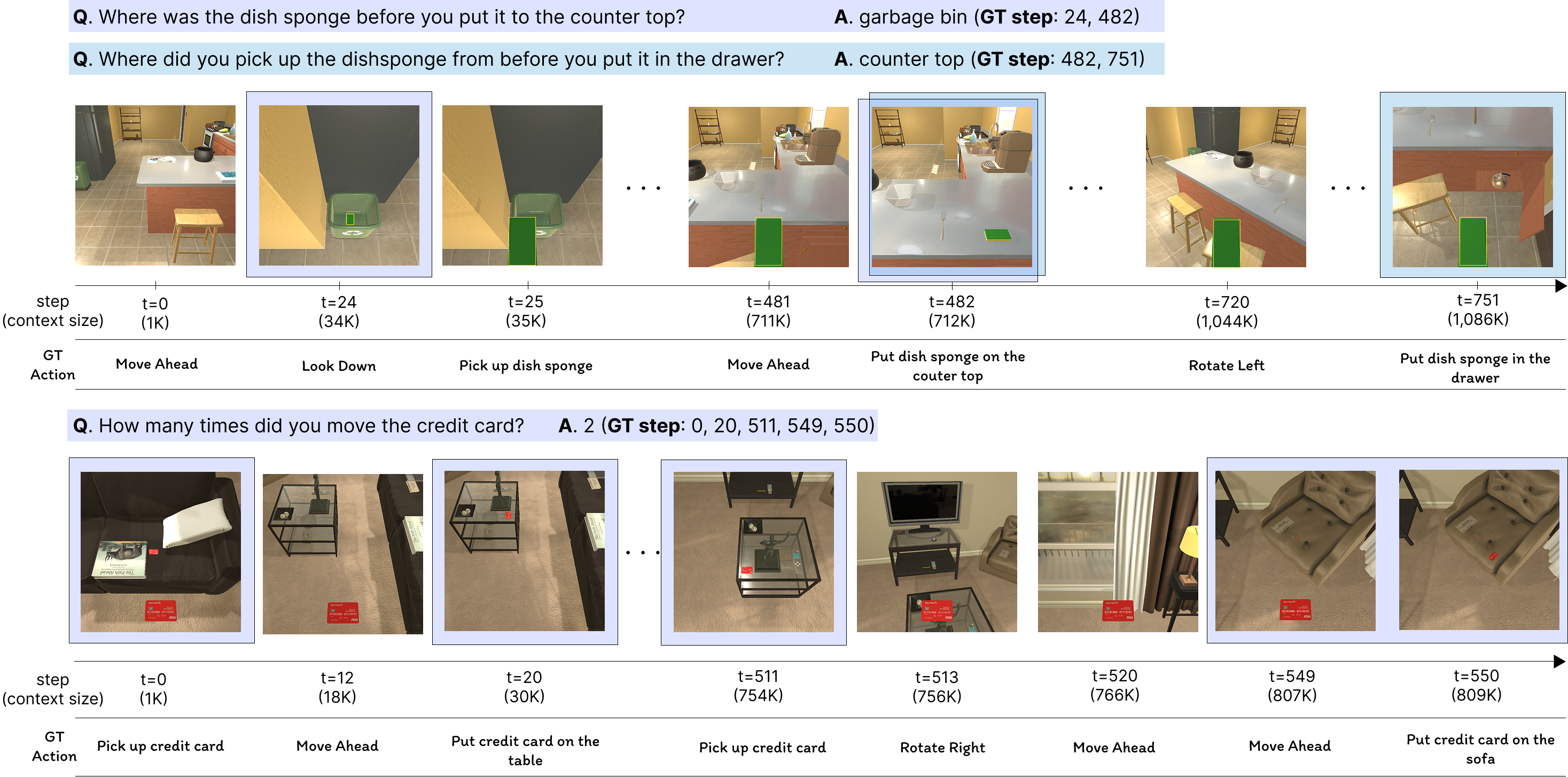}
  \label{fig:example_NiSSSEH}
}
\caption{Example of N(s)iEH task and Ground-truth steps.}
\vspace{-15pt}
\label{fig:NiEH}
\end{figure*}

\ourname features with a generation framework for synthesizing long trajectories to train and evaluate AI agents in long-horizon embodied tasks.
We build \ourname upon AI2-THOR \citep{kolve2017ai2} simulator, an interactive 3D environment for embodied AI research that supports diverse scenes, objects, and agent actions.
\ourname enables the creation of unlimited trajectories with arbitrary length, and provides an evaluation setup where agents can interact dynamically with the environment during both training and testing.
This supports both offline learning by producing large-scale datasets, and online learning through direct agent-environment interaction.

Each trajectory generated by \ourname consists of multiple task goals, such as ``Put a clean sponge on a metal rack'' and ``Pick up the apple and place it on the microwave'', requiring grounded understanding and action to achieve the goal.
At the end of each trajectory, the agent is assigned a synthetic long-horizon task that requires reasoning over entities encountered at distant time steps.
For the example in Figure~\ref{fig:fig1}, the long-horizon task (Sub-goal \#23) at step $t=670$, ``Put the tomato on the counter top'', depends on observations made far earlier: the tomato at $t=16$ and the counter top at $t=560$.

Our generation framework can generate unlimited tasks, the trajectories can be exceptionally long, exceeding 1M context tokens or beyond when the trajectory is processed with LLMs.
Successfully completing this task requires the agent to (1) memorize and integrate key environmental information over hundreds of steps, and (2) plan actions based on dependencies that are separated in time, demonstrating the need for long-context reasoning and robust spatio-temporal memory.

\subsection{Static Evaluation: Needle(s) in the Embodied Haystack}

We first introduce a novel task in the form of a static evaluation: Needle(s) in the Embodied Haystack (NiEH).
NiEH is designed to evaluate an agent’s ability to recall and reason about environmental states encountered throughout a trajectory.
Unlike traditional embodied QA tasks that focus primarily on visual understanding of a single image, NiEH emphasizes reasoning about environmental changes over time, requiring agents to interpret and integrate sequences of multimodal observations.

Figure~\ref{fig:NiEH} presents examples of the two NiEH task types.
In the single-evidence setting, a question is answerable based on a single observation step; in the multi-evidence setting, multiple temporally distant steps must be combined to answer the question.
The NiEH testset includes diverse question types, such as binary (``yes'' or ``no''), ``what''-, ``where''-, and ``how many''-style questions.
These questions span a broad range of difficulty, from simple memory recall (similar to the Needle in a Haystack paradigm \citep{liu-etal-2024-lost}) to complex queries that requiring multi-step reasoning across temporally and spatially distributed evidence.

\textbf{Testset Construction.}
We first replay the generated trajectories to collect egocentric views and record object interactions directly from the simulator's metadata.
Using rule-based templates, we generate QA pairs (e.g., ``Q. What object did you slice? A. \texttt{\{object\}}'') and sample them based on frequency to ensure diversity.
To ensure valid answerability, we cross-validate questions against ground-truth images using an ensemble of four multimodal LLMs: LLaVA-OneVision, Qwen2.5-VL, Deepseek-VL, and Pixtral \citep{li2024llava,bai2025qwen25vltechnicalreport,lu2024deepseekvlrealworldvisionlanguageunderstanding,agrawal2024pixtral12b}.
We filter out questions that none of these models can successfully answer.
Full details on generation rules and validation are provided in Appendix \ref{sec:nieh_construction}.

\textbf{Challenges in NiEH. }
NiEH introduces two key challenges for current models.
First, questions require reasoning over multiple temporally distant events.
As shown in Figure \ref{fig:NiEH}(b), the agent moves a \texttt{dish sponge} from the garbage bin at $t=24$, then to the counter top, and later places it into a drawer at $t=751$.
A question such as \textit{``Where was the dish sponge before you put it on the counter top?''} requires the model to recall and chain together multiple actions and locations across hundreds of steps.
Second, some questions demand aggregating sparse and temporally scattered evidence from long trajectories. 
In the second example in Figure \ref{fig:NiEH}(b), answering \textit{``How many times did you move the credit card?''} requires the model to track and count all relevant actions occurring from the beginning to the end of the episode.
These tasks necessitate robust long-horizon reasoning across both time and modalities in complex embodied environments.

\subsection{Constructing Long-horizon Trajectories for Interactive Evaluations}

With \ourname's generation framework, we synthesize long-horizon trajectories to construct training, validation, and test sets for offline learning and evaluation.
Our approach builds upon a planner-based method \citep{kolve2017ai2}, in which we sequentially concatenate multiple single-task demonstrations into a extended trajectory, while maintaining consistency in object states and agent interactions throughout.
To generate each trajectory, we first sample a task type from one of seven predefined templates (e.g., ``pick two objects and place'', ``pick and place with movable receptacle''), and objects.
Based on the sampled task and objects, we use a classical task planner that operates on domains specified in the Planning Domain Definition Language (PDDL) \citep{McDermott1998PDDLthePD} to generate ground-truth action sequences.
Only successful rollouts (re-simulated without failure) are retained, ensuring the reproducibility and reliability.
We then concatenate these successful demonstrations to construct long-horizon sequences that span hundreds of steps.
For the final goal, the involved objects are sampled exclusively from those seen during the early 20\% and the final 20\% of the trajectory.
This enforces a long-term temporal dependency between two objects that must be jointly referenced to complete the final task.
Details on task types and a pseudo-algorithm for the generation process are provided in Appendix \ref{sec:construct_long_traj}.

\begin{table}
\small
\caption{Dataset statistics. Token lengths are calculated using the LLaVA-OneVision tokenizer. See Appendix \ref{sec:more_statistics} for detailed statistics.}
\centering
\small
\begin{tabular}{lrrrr}
\toprule
\multicolumn{2}{l}{\textbf{NiEH testset}} & Single-clue & Multi-clue \\
\midrule
\multicolumn{2}{l}{\# of question-answer pair} & 829 & 474 \\
\midrule
\textbf{Trajectory} & \textbf{Train} & \textbf{Dev} & \textbf{Test} \\
\midrule
\# trajectory     &  2,456 & 125 & 225 \\
\# avg/max subgoals   & 14/30  & 16/24 & 18/33 \\
\# avg/max steps  & 405/654 & 613/890 & 627/952 \\
\# avg token length   & 602K  & 880K & 912K \\
\# max token length   & 954K  & 1.2M & 1.3M \\
\bottomrule
\end{tabular}
\vspace{-8pt}
\label{tab:stat}
\end{table}

\section{Architectures and Training Considerations for Long-Horizon Embodied Reasoning}

\begin{figure*}[t]
\centering
\includegraphics[width=1.0\textwidth]{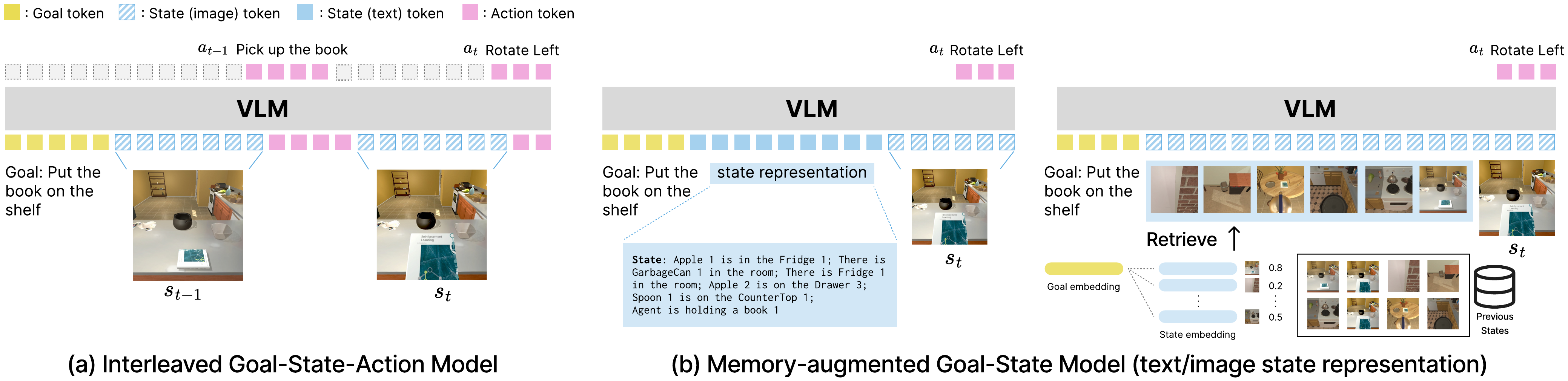}
\caption{
\textbf{Overview of Policy Architectures.}
{(a) Interleaved Goal-State-Action Model:} Encodes the full history as a single continuous sequence of goal, state, and action tokens.
{(b) Memory-augmented Goal-State Model:} Retrieves relevant past information to provide long-term context without processing the entire trajectory.
}
\vspace{-10pt}
\label{fig:archi}
\end{figure*}

Long-horizon embodied tasks pose significant challenges due to the need to interpret multimodal inputs (vision, language) and produce coherent sequences of actions over hundreds of interactive steps.
However, existing approaches either use separate encoders for vision, language, and action modules \citep{shridhar2020alfred,hafner2025trainingagentsinsidescalable}, or focus on short-horizon in constrained environments \citep{brohan2023rt1roboticstransformerrealworld}, where decisions depend only on the most recent observation and a single instruction.
While recent VLMs \cite{NEURIPS2023_6dcf277e,lu2024deepseekvlrealworldvisionlanguageunderstanding,bai2025qwen25vltechnicalreport} show strong multimodal reasoning abilities, these models are not directly suited to long-horizon embodied tasks, which are interaction-dominant, involving hundreds of egocentric frames and continuous vision-language-action sequences.
Moreover, many state-of-the-art proprietary models are only accessible via paid, rate-limited APIs, which makes large-scale, real-time experimentation in embodied settings more difficult and costly in practice \cite{li2024embodied}.
In this work, we explore two distinct policy architectures for leveraging a VLM as a unified embodied agent: (1) Interleaved Goal-State-Action Model and (2) Memory-Augmented Goal-State Model.

\textbf{Interleaved Goal-State-Action Model} treats the entire perception–action stream as a single unified sequence.
At each timestep, the input consists of the task goal followed by all past visual observations and actions in an interleaved order $(g_0, s_0, a_0, …, s_t, a_t)$.
This design enables the VLM backbone to jointly reason over multimodal information and autoregressively predict the next action conditioned on the full trajectory history.
By leveraging interleaved modeling, our architecture supports coherent decision-making over temporally distant information while maintaining grounded behavior in dynamic settings.

\textbf{Memory-Augmented Goal-State Model} incorporates compact history representations into the input context.
Instead of supplying the full trajectory, the model conditions on either
(1) textual summaries of past states and environment information (Figure~\ref{fig:archi}-(b)-left), or
(2) a retrieved set of relevant historical images selected via embedding similarity (Figure~\ref{fig:archi}-(b)-right).
These memory tokens provide distilled representations of long-term context, allowing the model to focus on the current observation while still retaining access to key historical information.
By combining memory tokens with the current state, the model can maintain semantic grounding over long horizons without exceeding context-length constraints.

\textbf{Context Extension and Context Parallelism}.
Given the limitations in context length of most VLMs, using off-the-shelf models is insufficient for processing long inputs such as those exceeding 1M tokens, particularly in our interleaved Goal-State-Action setting.
We explore various long-context extension techniques that allows the model to generalize to longer input sequences without retraining from scratch.
Specifically, we consider:
\textbf{Linear Interpolation}~\cite{chen2023extendingcontextwindowlarge}: Rescales input positions to fit within the pretrained RoPE range by linearly interpolating positional indices;
\textbf{Dynamic Scaling}~\cite{chen2023extendingcontextwindowlarge}: Adapts RoPE frequencies at runtime based on the input sequence length, using a linear rescaling to maintain consistent positional encoding behavior across varying lengths;
\textbf{YaRN}~\cite{peng2024yarn}: Dynamically interpolates attention frequencies during inference, balancing between pretrained and extrapolated positional regimes;
\textbf{LongRoPE}~\cite{ding2024longropeextendingllmcontext}: Augments RoPE with specially designed extrapolation functions, enabling robust generalization to long sequences without degrading attention quality.
We apply these techniques during fine-tuning, at inference time, or both.

We also incorporate {Context Parallelism} built upon Ring Attention \citep{liu2023ring}, which cyclically exchanges key-value shards across devices to compute full attention with significantly reduced memory overhead.
We apply these techniques during fine-tuning, inference, or both as essential tools for handling long-context inputs and evaluate how effectively they improve practical performance when interleaved trajectories become substantially long.

\section{Experiments}

\begin{figure*}[t]
\centering
\includegraphics[width=1.0\textwidth]{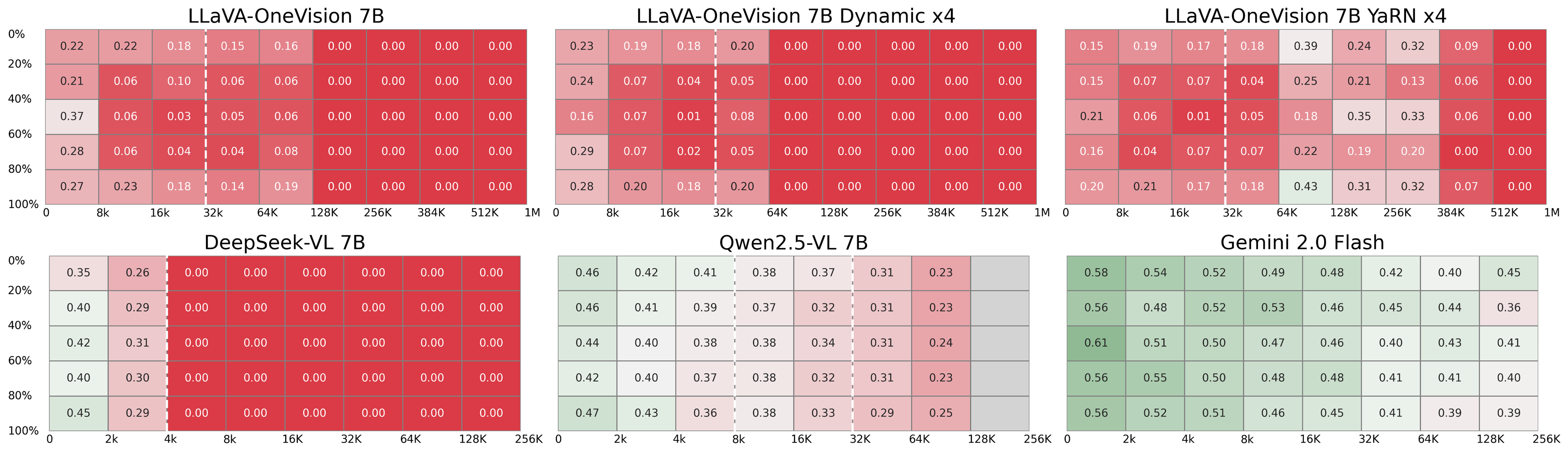}
\caption{
\textbf{Results of Needle(s) in the Embodied Haystack.}
The x-axis denotes the input context length (tokens), and the y-axis denotes the needle depth (percentage position of the evidence within the trajectory).
The white dashed line denotes the maximum pre-training context length of each model.
Qwen2.5-VL was initially pre-trained with an 8K token context window and incrementally scaled up to 32K tokens in subsequent stages \citep{bai2025qwen25vltechnicalreport}.
The gray area indicates contexts not applicable (N/A) due to the model's smaller image token size which limits sequences to under 128K tokens.
Context Parallelism is applied to all experiments with the context sizes over 384K.
Full results are provided in Appendix \ref{sec:more_result_nieh}.
}
\label{fig:results_combined_main}
\end{figure*}

\begin{table*}[t]
\centering
\small
\caption{
\textbf{Architectural Comparison.} QA performance on N(s)iEH across different architectures in a setting where the \textit{full trajectory context is available}.
LLaVA-OV model fails when provided with the full trajectory as input ($>$1M tokens).
GPT-4o is used as an LLM judge for evaluation.
For Memory-Augmented approaches, we use a CLIP encoder \citep{radford2021learning} to compute embedding similarity between goal descriptions and historical state observations, selecting the Top-10 and Top-20 most relevant images.
}
\resizebox{\linewidth}{!}{
\begin{tabular}{lcccccc}
\toprule
& \multicolumn{3}{c}{\textbf{NiEH} (Single-evidence)} &
  \multicolumn{3}{c}{\textbf{NsiEH} (Multi-evidence)} \\
\cmidrule(lr){2-4} \cmidrule(lr){5-7}
& LLaVA-OV & Qwen2.5-VL & Gemini 2.0 Flash
& LLaVA-OV & Qwen2.5-VL & Gemini 2.0 Flash \\
\midrule

(a) Full trajectory (Image-only)
& 0.0 & 44.67 & 35.44
& 0.0 & 19.34 & 37.08 \\

(b) Interleaved Goal-State-Action
& 0.0 & 51.64 & \textbf{83.04}
& 0.0 & 43.24 & \textbf{62.72} \\

(c) Memory-Augmented (Text)
& 51.33 & 54.38 & 56.21
& 53.83 & 51.33 & 40.99 \\

(d) Memory-Augmented (Image, Top-10)
& 35.80 & 33.99 & 25.18
& 22.15 & 18.63 & 21.18 \\

(e) Memory-Augmented (Image, Top-20)
& 37.36 & 42.68 & 28.16
& 23.80 & 21.46 & 30.25 \\
\bottomrule
\end{tabular}
}
\vspace{-10pt}
\label{tab:nieh-full}
\end{table*}

\subsection{Static Evaluation: Needle(s) in the Embodied Haystack}
We first evaluate model performance on the Needle in the Embodied Haystack (NiEH) and Needle\textbf{s} in the Embodied Haystack (NsiEH) tasks, which test an agent's ability to retrieve and reason over sparse evidence scattered throughout long embodied trajectories.

\textbf{Building a Embodied Haystack.} Unlike the traditional Needle in the Haystack setup, which inserts a target sentence into a long text corpus like a book, we use the entire embodied trajectory as the input context.
To simulate different reasoning depths, we crop the input sequence either from the beginning or the end based on the GT image's position.
In the NsiEH task, where multiple evidences are scattered throughout the trajectory, we fill the context with intermediate steps in between the GT steps keeping their temporal order.

\textbf{Results.} 
Figure \ref{fig:results_combined_main} (top row) shows the performance of LLaVA-OneVision (OV) 7B across different context extension methods.
While Dynamic scaling struggles with contexts beyond 64K tokens, YaRN x4 improves the baseline, handling sequences up to 384K tokens.
The second row compares different VLMs: DeepSeek-VL, Qwen2.5-VL, and Gemini 2.0 Flash.
Each model processes images into tokens differently--DeepSeek (576 tokens/image), Qwen2.5-VL (121 tokens/image), and Gemini 2.0 Flash (258 tokens/image)--which consequently impacts maximum context lengths when transforming N(s)iEH sequences into tokenized inputs (DeepSeek-VL: 512K, Qwen2.5-VL: 128K, Gemini 2.0 Flash: 256K).
DeepSeek-VL fails immediately beyond its 4K pre-training limit.
Qwen2.5-VL outperforms DeepSeek but shows clear degradation beyond its 32K pre-training limit.
Gemini 2.0 Flash performs robustly up to 32K tokens but performance drops below 40\% at 256K, indicating that complex long-range multimodal reasoning remains a challenge even for state-of-the-art models.
See Appendix \ref{sec:more_result_nieh} for more results.

\textbf{Architectural Comparison.}
Table \ref{tab:nieh-full} presents N(s)iEH task performance across different architectures in a setting where \textit{full trajectory context is available}.
LLaVA-OV fails to process these full trajectories ($>$1M tokens), exposing an inherent limitation in its ability to handle long contexts.
In contrast, Gemini 2.0 Flash effectively leverages this temporal context.
For Memory-Augmented approaches, textual memory demonstrates significantly higher performance than image-based memory,
as textual state representations provide denser information within a shorter context size.
Notably, we observe that Gemini 2.0 Flash tends to hallucinate object locations in image-only settings when language grounding is not provided (Table \ref{tab:nieh-full}-(a), (d), and (e)).

\begin{figure*}[t]
\centering
\includegraphics[width=1.0\textwidth]{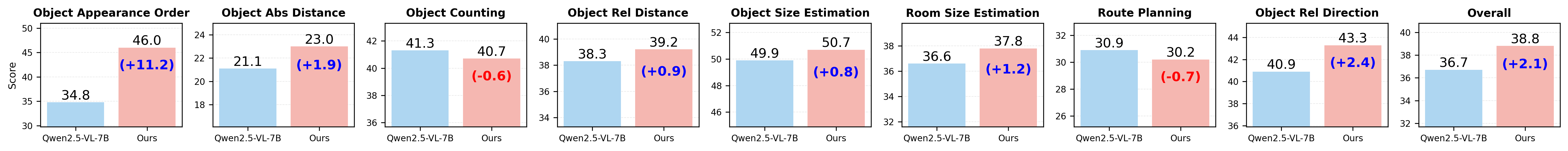}
\caption{
\textbf{Sim2Real Transfer on VSI-Bench.}
We compare Qwen2.5-VL 7B (Base) against the same model fine-tuned on 226K synthetic NiEH QA pairs.
Our model achieves a +11.2\% in \textit{Object Appearance Order} and a +2.1\% improvement overall, demonstrating effective transfer to photorealistic environments.
}
\vspace{-10pt}
\label{fig:vsibench}
\end{figure*}

\textbf{Single vs Multi-evidence Reasoning.}
Comparing NiEH to the more challenging NsiEH task, we observe a significant performance drop in the multi-evidence setting.
This is especially pronounced for questions involving sparse or distant evidence (e.g., \textit{``Where was the Mug before you put it on the CounterTop?''}) or questions requiring the aggregation of multiple evidences (e.g., \textit{``How many times did you move the Apple?''}), as shown in Figure \ref{fig:NiEH}(b).
These results demonstrate that our NiEH and NsiEH tasks pose a substantial challenge to current long-context models and success requires both fine-grained temporal memory and multi-evidence reasoning across extended interactions.

\subsection{Online Evaluation: Long-horizon Tasks}
We conduct an \textit{online} evaluation using the AI2THOR simulator.
We evaluate models on the final long-horizon task of each test trajectory (e.g., Sub-goal \#23 in Figure 1), which requires the agent to utilize history from the preceding hundreds of steps.
We fine-tune both LLaVA-OV and Qwen2.5-VL on our training set.
Detailed training specifics are available in Appendix \ref{sec:training_detail}.

\textbf{Results.} Table \ref{tab:online_eval} presents the success rates from our interactive online evaluation.
LLaVA-OV failed to complete any long-horizon tasks in this setting ($>$1M tokens); therefore, we report results only for the Qwen2.5-VL model in Table \ref{tab:online_eval}.
We observe that interleaved models consistently outperform Memory-Augmented approaches, suggesting that processing the continuous history is more effective than retrieval-based methods for grounded reasoning.
Training with longer contexts leads to significant gains: the 128K context model achieves a 9.2\% success rate, confirming that access to extended history is crucial for long-horizon performance.
Incorporating long-context extension techniques further boosts performance.
YaRN (x4) achieves the highest total success rate of {11.5\%}, providing a +2.3\% gain over the base interleaved model.
However, the \texttt{Put} task remains unsolved across all methods, as \texttt{Put} task comes after \texttt{Go to} and \texttt{Pick up}, and the agent rarely reaches this final stage due to failures in the earlier steps.
Evaluation results for full-episode execution are available in Appendix \ref{sec:online_eval_full_traj}.
Overall, these results demonstrate that online evaluation for long-horizon tasks remains a significant open challenge.

\begin{table}[t]
\centering
\caption{
\textbf{Online Evaluation. }
Success Rates (\%) for the long-horizon tasks.
32K and 128K denote the context size used during fine-tuning.
}
\label{tab:online_eval}
\resizebox{\linewidth}{!}{
\begin{tabular}{l|ccc|c}
\toprule
\multicolumn{1}{c}{\textbf{Method}} & \texttt{Go to} & \texttt{Pick up} & \texttt{Put} & \textbf{Total} \\
\midrule
Memory-Augmented (Text) & 10.8 & 1.5 & 0.0 & 6.1 \\
Memory-Augmented (Image, Top-20) & 6.2 & 0.0 & 0.0 & 3.1 \\
\midrule
Interleaved Goal-State-Action (32K) & 11.3 & 1.5 & 0.0 & 6.4 \\
Interleaved Goal-State-Action (128K) & 18.5 & 2.3 & 0.0 & 9.2 \\
+ Dynamic Scaling & 20.0 & 2.3 & 0.0 & 9.9 \\
\textbf{+ YaRN (x4)} & \textbf{23.1} & \textbf{2.6} & 0.0 & \textbf{11.5} \\
\bottomrule
\end{tabular}
}
\vspace{-10pt}
\end{table}

\subsection{Demonstrations: Sim2Real Transfer and Low-level Control}

Finally, we demonstrate the versatility of \ourname through Sim2Real transfer and integration with low-level control policies.

\textbf{Sim2Real Transfer.}
We show the transferability of our framework. We used the NiEH pipeline to generate 226K synthetic QA pairs from the training set and fine-tuned the Qwen2.5-VL model.
We evaluated the model on VSI-Bench \cite{Yang2024ThinkingIS}, a \textit{photorealistic} benchmark for visual-spatial reasoning.
As shown in Figure~\ref{fig:vsibench}, our model achieves +11.2\% improvement on the \textit{Object Appearance Order} task, confirming that temporal reasoning skills learned in simulation transfer to real-world images.
Overall, we observe a +2.1\% performance gain with negligible degradation in other spatial tasks.

\textbf{Low-level Manipulation.}
We integrated \ourname with ManipulaTHOR \cite{Ehsani2021ManipulaTHORAF} to enable fine-grained, low-level robot arm manipulation control.
Table \ref{tab:low_level_success} reports the success rates on these tasks using OpenVLA-7B \citep{kim24openvla} and SpatialVLA-4B \citep{qu2025spatialvlaexploringspatialrepresentations}.
Ego-centric views are provided as image input, and the models generate the next action $(\Delta x, \Delta y, \Delta z)$ to control the robot arm.
Overall performance remains low, with both models struggling on manipulation tasks.
We attribute these results primarily to two factors: (1) ManipulaTHOR’s egocentric view, which provides nearly identical images for fine-grained arm movements $(\Delta x, \Delta y, \Delta z)$ smaller than 0.01, making it difficult for VLA models to perceive subtle adjustments;
and (2) discrepancies between the robot arm model in ManipulaTHOR and those assumed by the VLA models, causing many predicted actions to fail in the physical simulation.
We provide more results and analysis in Appendix \ref{sec:more_result_low_level}.

\begin{table}[t]
\small
\centering
\caption{\textbf{Low-level Manipulation Task.} Success rates (\%) of OpenVLA-7B and SpatialVLA-4B on manipulation tasks in ManipulaTHOR.}
\label{tab:low_level_success}
\begin{tabular}{lcc}
\toprule
\multicolumn{1}{c}{\textbf{Task}} & \textbf{OpenVLA-7B} & \textbf{SpatialVLA-4B} \\
\midrule
\texttt{Pick up} & 16.52 & 18.70 \\
\texttt{Put}     & 2.08 & 1.19 \\
\bottomrule
\end{tabular}
\vspace{-10pt}
\end{table}

\section{Conclusion}

We presented \ourname, a new framework for long-horizon embodied tasks designed to advance long-context understanding in embodied AI.
Our framework enables scalable synthesis of long, complex trajectories paired with ground-truth action sequences, and supports both offline training and online interaction with the environment.
As part of this framework, we introduced a novel embodied QA benchmark, {Needle(s) in the Embodied Haystack}, that challenges agents to reason over sparse, temporally distant visual evidence embedded within extended trajectories.
To equip models for this setting, we explored policy architectures including interleaved Goal-State-Action and Memory-Augmented Goal-State modelings, along with context extension techniques and efficient fine-tuning via Context Parallelism.
Our experiments demonstrate that while inference-time extension provides gains, access to longer contexts during training is critical for robust performance.
Furthermore, we validated the versatility of \ourname through successful Sim2Real transfer to photorealistic benchmarks and integration with low-level manipulation controls.
Despite these advances, our results highlight that long-horizon reasoning remains an open challenge, with current state-of-the-art VLMs still struggling to reason over extreme context lengths.
We hope our framework encourages further research into the next generation of embodied agents capable of robust long-horizon reasoning and planning.

\section*{Impact Statement}

This paper introduces a framework for training and evaluating long-horizon embodied agents, with novel benchmarks, architectural explorations, and training methodologies.
We view our framework as a platform to teach embodied agents how to reason over long temporal horizons and execute complex, multi-step instructions in dynamic environments.

The most relevant downstream task for this work is long-horizon embodied planning and robotic manipulation.
As our dataset is synthesized within the AI2-THOR simulator, there is a risk of domain gap and limited diversity in scene configurations, which may result in performance biases when applied to unstructured real-world environments.

\bibliography{example_paper}
\bibliographystyle{icml2026}

\newpage
\appendix
\onecolumn

\section{Dataset Construction}
\label{sec:dataset_construction}

\subsection{Building the Needle(s) in the Embodied Haystack Benchmark}
\label{sec:nieh_construction}

We construct the Needle(s) in the Embodied Haystack benchmark in three stages:
1) Trajectory Replay and Metadata Collection;
2) Rule-Based QA Generation; and
3) Cross-validation with Multimodal LLMs.
The following sections provide detailed descriptions of each step.

\subsubsection{Trajectory Replay and Metadata Collection}

We first replay 225 test trajectories generated by \ourname, logging both visual observations (agent's egocentric views) and structured metadata at each timestep.
For every step, we store the list of visible objects, agent-inventory items, openable containers, and their contents from the simulator.
This produces a fine-grained interaction log that captures grounded scene dynamics over time.

An example of the collected metadata at a single timestep is shown below:

\vspace{0.5em}
\begin{tcolorbox}[colback=gray!0!white, colframe=black!70, title=Example of metadata entry]
\footnotesize
\begin{verbatim}
{
    "img_idx": 2,
    "img_filename": "000000002.png",
    "step": 1,
    "object_log": {
        "visible": ["Shelf", "Vase", "Book"],
        "pickupable": ["Vase", "Book"],
        "isOpen": [],
        "inven_obj": [],
        "receptacles": ["Shelf"],
        "recep_objs": {
            "Shelf": ["Vase", "Book"]
        }
    }
}
\end{verbatim}
\end{tcolorbox}
\vspace{0.5em}

Each metadata entry corresponds to a low-level action step and provides the semantic state of the scene, enabling the construction of temporally grounded QA instances in later stages.

\subsubsection{Rule-Based QA Generation}
To construct the QA set, we apply rule-based generation templates to each trajectory using its sequence of low-level actions and associated metadata.
The QA generation process involves parsing the agent's interactions with objects, containers, and the environment, and applying a set of handcrafted rules to synthesize grounded questions.

Our QA generation logic covers a diverse range of question types, including object presence, object state, location tracking, slicing actions, container content reasoning, and action counting.
For instance, if an object is seen for the first time at a particular step, a presence question such as ``Is there any \texttt{apple} in this room?'' is generated.
Similarly, after a \texttt{PutObject} action, location-based questions like ``Where was the \texttt{apple} before you put it to the \texttt{microwave}?'' are produced.
When slicing actions happen, we create questions about the object being sliced and other nearby items (e.g., “What objects were in the \texttt{Fridge} when you sliced the \texttt{apple}?”).
Then, we sample questions based on the frequency to ensure diversity across object types, and annotate the GT answer steps using the replay logs.
Table~\ref{tab:qa-types-templates} summarizes the types of questions generated, and corresponding trigger conditions and example templates.

\begin{table}[h]
\centering
\footnotesize
\caption{QA types, trigger conditions, and corresponding question templates used in rule-based generation.}
\label{tab:qa-types-templates}
\begin{tabular}{@{}p{3.1cm} p{3.2cm} p{6.6cm}@{}}
\toprule
\multicolumn{1}{c}{\textbf{QA Type}} & \multicolumn{1}{c}{\textbf{Trigger Condition}} & \multicolumn{1}{c}{\textbf{Example Template(s)}} \\
\midrule
object presence (Yes/No) &
object appears visibly in the trajectory &
\texttt{Is there any \{obj\} in this room?} \newline
\texttt{Have you seen a/an \{obj\}?} \\
open state questions &
container marked as open in metadata &
\texttt{Was \{container\} open?} \\
object location tracing &
sequences of \texttt{Pickup} and \texttt{PutObject} actions &
\texttt{Where was \{obj\} before you put it to \{container\}?} \newline
\texttt{Where did you move the \{obj\} from the \{container\}?} \newline
\texttt{Where is \{obj\} now?} \\
slicing-based questions &
\texttt{SliceObject} action detected in trajectory &
\texttt{What did you slice?} \newline
\texttt{What objects were in/on the \{container\} when you slice the \{obj\}?} \\
container content &
container visibility with non-empty contents &
\texttt{What objects were in/on the \{container\}?} \newline
\texttt{What object did you put in/on the \{container\}?} \\
put action questions &
unique \texttt{PutObject} action for a container &
\texttt{What object did you put in/on the \{container\}?} \\
final object state &
final location of an object at episode end &
\texttt{Is \{obj\} in/on the \{container\}?} \newline
\texttt{What objects are in/on the \{container\}?} \newline
\texttt{How many objects were in/on the \{container\}?} \\
movement counting &
object picked up more than once &
\texttt{How many times did you move \{obj\}?} \\
\bottomrule
\end{tabular}
\end{table}

\subsubsection{Cross-validation with Multimodal LLMs}

To ensure the answerability and clarity of the generated QA pairs, we perform cross-validation using four powerful multimodal LLMs: LLaVA-OneVision 7B~\citep{li2024llava}, Qwen2.5-VL 7B~\citep{bai2025qwen25vltechnicalreport}, Deepseek-VL 7B~\citep{lu2024deepseekvlrealworldvisionlanguageunderstanding}, and Pixtral 12B~\citep{agrawal2024pixtral12b}.
Each model is prompted with the GT images corresponding to the annotated QA steps and asked to answer the associated questions.
Given their strong performance on standard visual QA tasks, we use these models to assess whether a question can be correctly answered or not.
We keep only the QA pairs that are correctly answered by at least one of the four models, and discard those that fail across all models.
This helps improve dataset quality and filtering out ambiguous or visually ungroundable questions.
Table \ref{tab:cross-validation} shows the accuracy of each model on the finalized QA set when evaluated with GT images.
Notably, even with access to GT images, all models struggle with questions requiring reasoning over three or more evidence steps.
To maintain the benchmark's difficulty and support evaluation of more capable models in future, we manually inspect the multi-clue questions and include those that are answerable.

\begin{table}[h]
\centering
\small
\caption{QA accuracy (\%) of multimodal LLMs on ground-truth images.}
\label{tab:cross-validation}
\begin{tabular}{l|c|rrr|r}
\toprule
\multicolumn{1}{c}{\textbf{Model}} & \textbf{Size} & \multicolumn{3}{c|}{\textbf{\# of clues (GT steps)}} & \textbf{Total} \\
 & & \textbf{1} & \textbf{2} & \textbf{$\geq$3} & \\
\midrule
LLaVA-OneVision & 7B & 86.61 & 68.55 & 23.74 & 71.15 \\
Qwen2.5-VL      & 7B & 85.94 & 89.83 & 64.40 & 82.20 \\
Deepseek-VL     & 7B & 81.56 & 39.14 &  22.57 & 62.88 \\
Pixtral         & 12B & 91.34 & 39.60 & 58.56 & 76.25 \\
\bottomrule
\end{tabular}
\end{table}

\subsection{Constructing Long-Horizon Trajectories}
\label{sec:construct_long_traj}

\begin{figure*}[h]
\centering
\includegraphics[width=1.0\textwidth]{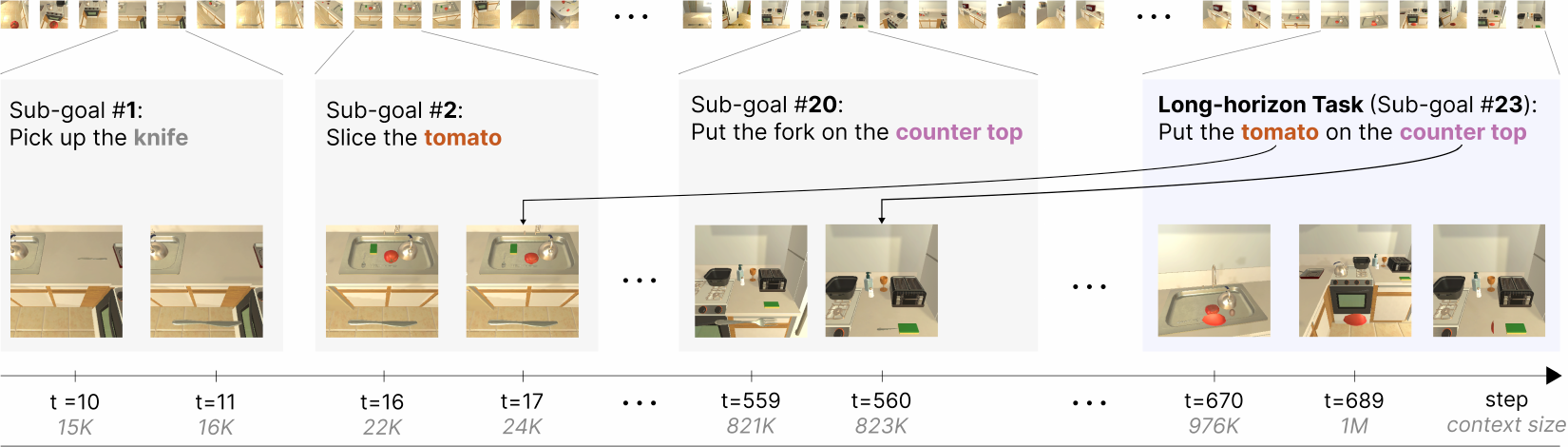}
\caption{
Example of the trajectory and a long-horizon embodied task generated from \ourname.
The final goal (``Put the tomato on the counter top'' at t=670) requires recalling both the tomato (seen at t=17) and the counter (seen at t=560) to solved the long-horizon task.
Context size refers to the input token length when converting the trajectory into the VLM input space.
}
\label{fig:fig1_old}
\end{figure*}

\begin{algorithm}[h]
\small
\caption{Construct Long-horizon Trajectory}
\label{alg:trajectory-generation}
\begin{algorithmic}[1]
    \STATE \textbf{Input:} Task Pool $\mathcal{T}$, max sub goals $N$
    \STATE \textbf{Output:} Long-horizon trajectory $\tau$

    \STATE Initialize empty trajectory $\tau \gets []$

    \WHILE{len($\tau$) $<$ $N$}
        \STATE Sample task template $g \sim \mathcal{T}$ and objects
        \STATE Plan action sequence $\pi_g$ by planner

        \IF{Simulate($\pi_g$) is successful}
            \STATE Append $\pi_g$ to trajectory: $\tau \gets \tau \mathbin{\|} \pi_g$
        \ELSE
            \STATE Discard and re-sample
        \ENDIF
    \ENDWHILE

    \STATE \COMMENT{Final sub-task with long-term object dependency}
    \STATE Sample $g_{\text{final}} \sim \mathcal{T}$ and objects in early 20\% and late 20\%
    \STATE Plan and simulate $\pi_{\text{final}}$ using restricted objects
    \IF{Simulate($\pi_{\text{final}}$) is successful}
        \STATE Append $\pi_{\text{final}}$ to trajectory: $\tau \gets \tau \mathbin{\|} \pi_{\text{final}}$
    \ELSE
        \STATE Repeat sampling until success
    \ENDIF

    \STATE \textbf{return} $\tau$
\end{algorithmic}
\end{algorithm}

Figure \ref{fig:fig1_old} shows an example of the trajectory and a long-horizon embodied task generated from \ourname.
To synthesize long-horizon trajectories, we construct each trajectory by sequentially chaining successful sub-tasks sampled from a predefined set of task templates. This process is illustrated in Algorithm \ref{alg:trajectory-generation}.
We begin by sampling a task template from a fixed task pool, which includes goal types such as \texttt{pick and place simple}, \texttt{pick two obj and place}, and \texttt{pick and place with movable recep}.
Each sampled template requires relevant objects in the scene (e.g., pickupable items, target receptacles), 
which are then used to define the goal for that task.

We use a classical task planner, which operates over PDDL-defined domains \citep{shridhar2020alfred}, to generate a low-level action sequence for the sampled goal, and simulate this plan in an interactive environment.
If the rollout fails (e.g., due to collisions, object occlusions, or unreachable conditions), we discard the sequence and re-sample from the task pool.
Otherwise, the successful rollout is retained and appended to the ongoing trajectory.

This sampling-execution loop is repeated until a long trajectory with a desired number of sub-goals is formed.
The resulting synthetic long-horizon trajectory consists of multiple sub-goals concatenated into a continuous sequence.
To induce long-term temporal dependencies, the final sub-task is constrained to involve only objects that appear in the early 20\% and late 20\% of the overall trajectory, requiring the agent to integrate temporally distant evidence to answer associated questions.

\subsection{Dataset Statistics}
\label{sec:more_statistics}

Table~\ref{tab:stat_appendix} summarizes the statistics for the NiEH test set and the generated long-horizon trajectories.
The training set consists of 2,456 trajectories with an average of 14 sub-goals and 405 steps per episode.
Token lengths are calculated using the LLaVA-OneVision tokenizer.
With an average input length of 602K tokens, the entire training dataset amounts to approximately 1.48B tokens (2,456 $\times$ 602K), providing a substantial corpus for learning long-horizon temporal dependencies.

\begin{table}
\small
\caption{Dataset statistics}.
\centering
\small
\begin{tabular}{lrrrr}
\toprule
\multicolumn{2}{l}{\textbf{NiEH testset}} & Single-clue & Multi-clue \\
\midrule
\multicolumn{2}{l}{\# of question-answer pair} & 829 & 474 \\
\midrule
\textbf{Trajectory} & \textbf{Train} & \textbf{Dev} & \textbf{Test} \\
\midrule
\# trajectory     &  2,456 & 125 & 225 \\
\# avg subgoals   & 14  & 16 & 18 \\
\# max subgoals   & 30  & 24 & 33 \\
\# avg steps  & 405 & 613 & 627 \\
\# max steps & 654  & 890 & 952 \\
\# avg token length   & 602K  & 880K & 912K \\
\# max token length   & 954K  & 1.2M & 1.3M \\
\# avg interaction with objects per episode & 31 & 36 & 37 \\
\# max interaction with objects per episode & 82 & 62 & 74 \\
\# scene   & 82  & 30 & 39 \\
\bottomrule
\end{tabular}
\vspace{-8pt}
\label{tab:stat_appendix}
\end{table}

\section{Training Details}

\label{sec:training_detail}
We fine-tune the LLaVA-OneVision 7B and Qwen2.5-VL 7B model on our training set while freezing the vision encoder.
Training utilized 8 H100 GPUs with tensor \cite{shoeybi2020megatronlmtrainingmultibillionparameter} and pipeline parallelism \cite{huang2019gpipeefficienttraininggiant} for 32K context, and Context Parallelism for larger contexts (64K, 130K).
The model is trained using a next-action prediction objective, where only the action tokens are optimized, conditioned on the goal and state tokens.
Table \ref{tab:training-specs} summarizes the training specifications for different context lengths.
For 32K training, we apply tensor parallelism with a degree of 4 and pipeline parallelism with a degree of 2, utilizing 8 H100 GPUs in total.
Since pipeline parallelism requires the batch size to match the pipeline degree, we set the batch size to 2.
For longer context lengths, we use context parallelism: 8-way for 64K (on 8 GPUs) and 16-way for 130K (on 16 GPUs).
All models are fine-tuned for approximately 3 epochs with a learning rate of 1e-5, using the AdamW optimizer and a linear learning rate schedule with a 0.03 warmup ratio.

\begin{table}[h]
\centering
\footnotesize
\caption{Training specifications for different context lengths.}
\label{tab:training-specs}
\begin{tabular}{@{}cccc@{}}
\toprule
\textbf{Context Length} & \textbf{Parallelism} & \textbf{\# GPUs} & \textbf{Training Time} \\
\midrule
32K & Tensor (4) + Pipeline (2) & 8 & 160 hrs \\
64K & Context (8)              & 8 & 120 hrs \\
130K & Context (16)             & 16 & 134 hrs \\
\bottomrule
\end{tabular}
\vspace{0.5em}
\end{table}

\section{Additional Results}

\subsection{Static Evaluation: Needle(s) in the Emboided Haystack}
\label{sec:more_result_nieh}

\begin{figure*}[h]
\centering
\includegraphics[width=1.0\textwidth]{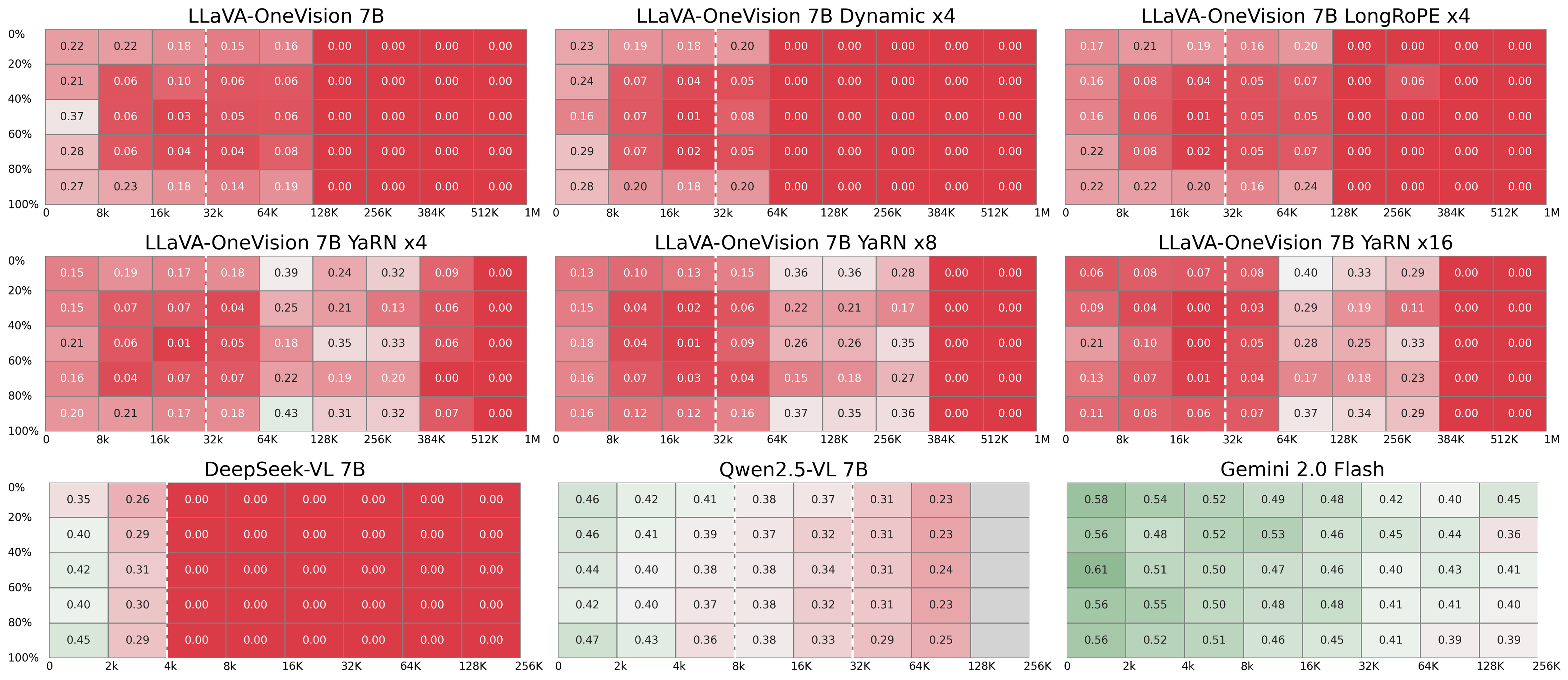}
\caption{
\textbf{Full results of Needle(s) in the Embodied Haystack.}
The x-axis denotes the input context length (tokens), and the y-axis denotes the needle depth.
The white dashed line denotes the maximum pre-training context length of each model.}
\label{fig:results_combined_all}
\end{figure*}

\begin{figure*}[h]
\centering
\includegraphics[width=1.0\textwidth]{figs/reward_1x4_.png}
\caption{
Agent's reward across different experimental configurations for high-level planning tasks. We compare (a) context extension methods at fixed scaling (x4), (b) varying YaRN scaling factors, and (c) fine-tuning with different context lengths using Context Parallelism. (d) summarizes the most effective strategies, highlighting that exposure to longer contexts during training significantly improves performance. Non-planner models cannot generate valid actions after around 250 steps ($\approx$376K in context size).
}
\label{fig:agent_reward}
\end{figure*}

Figure \ref{fig:results_combined_all} presents the performance of LLaVA-OneVision (OV) 7B \citep{li2024llava} model across various context extension methods.
Linear Interpolation, Dynamic Scaling, and LongRoPE scaling all struggled with very long contexts beyond 128K tokens (the results of Linear Interpolation are excluded from Figure since it fails at all examples).
YaRN consistently outperformed other methods, effectively handling contexts above 384K tokens, likely due to its architectural alignment with LLaVA-OV's Qwen2 LM backbone, which employs RoPE and YaRN scaling during pretraining.
YaRN performed best at moderate scaling factors (e.g., x4), however, further scaling to x8 and x16 did not yield additional gains.
In particular, x16 slightly improved performance in the 256K–384K token range but led to degradation notably at shorter context sizes ($<$64K), suggesting that excessive scaling may introduce instability and negatively impact performance.
Overall, all methods fail beyond 512K tokens, highlighting a need for improved long-context methods.
The third row includes additional results from DeepSeek-VL 7B \citep{lu2024deepseekvlrealworldvisionlanguageunderstanding}, Qwen2.5-VL 7B \citep{bai2025qwen25vltechnicalreport}, and Gemini 2.0 Flash\footnote{We used the \texttt{gemini-2.0-flash-001} version for all experiments.}.
Although Gemini is renowned for strong long-context multimodal reasoning (up to 3,000 image input), our results ($<$40\% with full trajectories at 256K) indicate substantial room for improvement.

\subsection{Online Evaluation: High-level Planning}
\label{sec:online_eval_full_traj}

\textbf{Plan-level Evaluation.}
We evaluate agent performance using a plan-level framework, where each plan corresponds to a short sequence of actions aimed at achieving a specific intermediate sub-goal (e.g., navigating to an object, placing an item). A trajectory is composed of multiple such plans, executed sequentially.
For the interactive evaluation, the agent is presented with the current plan’s goal along with the history of previous GT states and actions. Using this context, the agent predicts the next action and interacts step-by-step with the environment. The interaction continues until the current plan is either successfully completed or terminated due to failure (e.g., collisions or deadlocks).
After each plan, the context is reset to include the GT actions and states from the completed portion of the trajectory, and the agent proceeds to the next plan. This ensures that each plan is evaluated independently, conditioned only on the correct prior history.
The agent’s performance is measured via cumulative reward across all plans in the trajectory. Pseudocode for this evaluation procedure is provided in Algorithm~\ref{alg:plan-eval}.

\begin{algorithm}[h]
\small
\caption{Plan-Level Evaluation}
\label{alg:plan-eval}
\begin{algorithmic}[1]
\STATE \textbf{Input:} Trajectory $T = \{P_1, P_2, \ldots, P_N\}$, Agent policy $\pi$, Environment $\mathcal{E}$
\STATE \textbf{Initialize:} Reward $R \gets 0$
\STATE Initialize state and history with initial observation

\FOR{each plan $P_i$ in $T$}
    \STATE Initialize context with GT actions up to $P_{i-1}$
    \WHILE{not \textit{done} and not \textit{failure}}
        \STATE $a_t \gets \pi(\text{context})$
        \STATE $s_{t+1}, r_t, \textit{done}, \textit{failure} \gets \mathcal{E}.\text{step}(a_t)$
        \STATE Append $(a_t, s_{t+1})$ to context
        \STATE $R \gets R + r_t$
    \ENDWHILE
    \IF{\textit{failure}}
        \STATE Break evaluation
    \ENDIF
\ENDFOR

\STATE \STATE \textbf{return} Total accumulated reward $R$
\end{algorithmic}
\end{algorithm}

\textbf{Results and Discussion.} 
Figure \ref{fig:agent_reward} presents the accumulated rewards over time across four experimental configurations.
The Planner trajectory represents the performance upper bound.
Our analysis focuses on addressing the following key questions:

\textit{Q. Which context extension methods perform best?}
Figure \ref{fig:agent_reward}(a) compares different context extension methods at a fixed scaling factor of x4. Similar to the NiEH results, YaRN consistently achieves the highest performance showing very close performance to Planner.

\textit{Q. Does further scaling enhance performance?}
Figure \ref{fig:agent_reward}(b) explores YaRN scaling at different scaling factors (x4, x8, and x16). Interestingly, increasing the scaling factor beyond x4 does not significantly improve performance, indicating a diminishing return for larger scaling factors.

\textit{Q. Is fine-tuning on a dataset with long trajectories effective?}
Figure \ref{fig:agent_reward}(c) demonstrates the effectiveness of fine-tuning with Context Parallelism, enabling scaling of context lengths up to 64K and 130K tokens.
At a 130K context size, the model can learn sequences comprising approximately 86 steps, substantially longer compared to only 22 steps with a 32K context size.
This shows that exposure to longer context during training significantly enhances model performance, suggesting that incorporating more long-horizon data by \ourname could further improve model capabilities.
We note that context extension methods were not applied in this experiment.

\textit{Q. Does combining context extension methods during both training and inference provide additional benefits?}
Results of experiments with scaling at both training and evaluation (Figures \ref{fig:agent_reward}(d)) indicate that additional scaling at evaluation after fine-tuning with scaled RoPE provides no further performance improvement and may degrade performance at shorter context lengths ($\leq$300K tokens).

Based on these observations, we can conclude that
fine-tuning strategies are most effective when long-trajectory datasets are available.
In the absence of extensive training data, employing YaRN scaling at x4 yields performance comparable to the Planner upper-bound, particularly within context lengths under 200K tokens.

\subsection{Online Evaluation: Low-level Manipulation}
\label{sec:more_result_low_level}

We evaluate the ability of existing VLA models, OpenVLA-7B \citep{kim24openvla} and SpatialVLA-4B \citep{qu2025spatialvlaexploringspatialrepresentations}, to control robot arms on low-level manipulation tasks. The evaluation protocol follows the Algorithm~\ref{alg:plan-eval}, with the only difference being that low-level VLA models are used to directly execute \texttt{Pick up} and \texttt{Put} actions through arm control.
For the text input, we provide task-specific instructions (\texttt{Pick up} or \texttt{Put} object), since the existing models are only trained on single-task formulations.
For the image input, we use an ego-centric camera view.
Since these VLA models were originally trained on datasets with different viewing angles, there remains considerable room for improvement by incorporating additional camera perspectives during training and evaluation.

\begin{wrapfigure}{R}{0.32\textwidth}
    \centering
    \includegraphics[width=\linewidth]{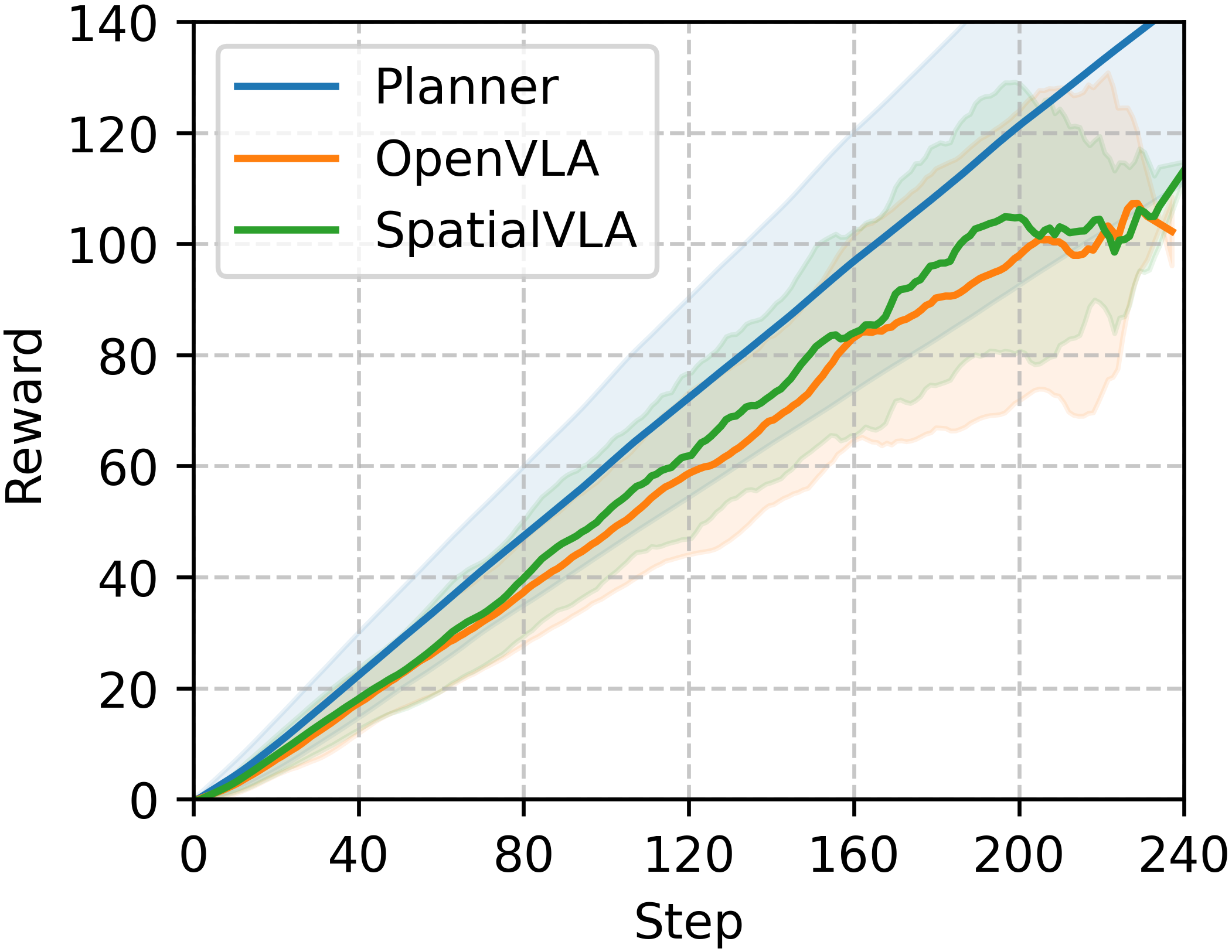}
    \caption{Agent's reward with low-level manipulation tasks. Low-level VLA models are used to control the robot arm for Pick-up and Put actions.}
    \label{fig:agent_reward_low_level}
\end{wrapfigure}

For the evaluation criterion, ManipulaTHOR implements a ``magnet sphere'' hand mechanism: if a pickupable object is within a specified radius of the agent’s hand when the \texttt{PickupObject} action is called, the object is successfully picked up.
Since neither OpenVLA nor SpatialVLA was trained in the AI2-THOR environment, we relax this success threshold by setting the radius to 0.4 to account for discrepancies between training and evaluation environments.
This loosened criterion ensures that small deviations in arm trajectories do not result in an immediate failure, thereby providing a fairer comparison of the models.

\textbf{Result.}
Figure~\ref{fig:agent_reward_low_level} presents accumulated rewards on low-level manipulation tasks.
Both models underperform relative to the Planner baseline, largely due to differences in robot arm configuration and the out-of-distribution nature of the visual inputs.
The results show that SpatialVLA achieves slightly higher and more stable rewards than OpenVLA across long sequences.

\end{document}